\documentclass[preprint,12pt]{elsarticle}

\usepackage{amsmath}
\usepackage{array}
\usepackage{booktabs}
\usepackage{bm}
\usepackage{graphicx}
\usepackage{multirow}
\usepackage{siunitx}
\usepackage{url}
\usepackage{comment}
\usepackage{subcaption}
\usepackage{pifont}

\usepackage[T1]{fontenc}
\newcommand{\xmark}{\ding{55}}%

\journal{}

\begin{document}

\begin{frontmatter}

\title{A simulation study of cluster search algorithms in data set generated by Gaussian mixture models}

\author[label1]{Ryosuke~Motegi\corref{cor1}}
\ead{t212d001@gunma-u.ac.jp}

\author[label2]{Yoichi~Seki}
\ead{sekiyoichi@gunma-u.ac.jp}

\cortext[cor1]{Corresponding author}

\affiliation[label1]{
            organization={Graduate School of Science and Technology, Gunma University},
            addressline={1-5-1 Tenjin},
            city={Kiryu},
            postcode={376-8515},
            state={Gunma},
            country={Japan}}

\affiliation[label2]{
            organization={Faculty of Informatics, Gunma University},
            addressline={1-5-1 Tenjin},
            city={Kiryu},
            postcode={376-8515},
            state={Gunma},
            country={Japan}}

\begin{abstract}
Determining the number of clusters is a fundamental issue in data clustering. Several algorithms have been proposed, including centroid-based algorithms using the Euclidean distance and model-based algorithms using a mixture of probability distributions. Among these, greedy algorithms for searching the number of clusters by repeatedly splitting or merging clusters have advantages in terms of computation time for problems with large sample sizes. However, studies comparing these methods in systematic evaluation experiments still need to be included. This study examines centroid- and model-based cluster search algorithms in various cases that Gaussian mixture models (GMMs) can generate. The cases are generated by combining five factors: dimensionality, sample size, the number of clusters, cluster overlap, and covariance type. The results show that some cluster-splitting criteria based on Euclidean distance make unreasonable decisions when clusters overlap. The results also show that model-based algorithms are insensitive to covariance type and cluster overlap compared to the centroid-based method if the sample size is sufficient. Our cluster search implementation codes are available at \url{https://github.com/lipryou/searchClustK}.

\end{abstract}


\end{frontmatter}

\section{Introduction}
The fundamental challenge in clustering is to find ``natural'' clusters in a data set. This challenge includes determining the number of clusters $K$ and choosing the distances or similarities when applying the clustering algorithm.

Centroid-based methods~\cite{centroid_based}, representing each cluster using centroids such as $K$-means, and model-based methods~\cite{bouveyron_celeux_murphy_raftery_2019}, which assume probability distributions for each cluster, are often used. The most used is the $K$-means method~\cite{EZUGWU2022104743}. With $K$-means, the number of clusters $K$ must be specified by the user. It is crucial to run multiple times with different $K$ to select a reasonable estimation. Although there are various validity indices for selecting $K$ in $K$-means~\cite{nbclust}, it depends on the circumstances in which indices should be used. In contrast, a model-based method has the advantage that an information criterion can objectively evaluate the fit of the probability model to the data set. Akaike information criterion (AIC)~\cite{AIC}, Bayesian information criterion (BIC)~\cite{BIC}, and minimum description length (MDL)~\cite{MDL} are commonly used as information criteria. The finite mixture model (FMM)~\cite{bouveyron_celeux_murphy_raftery_2019}, a probability model that mixes finite probability distributions, such as the Gaussian mixture model (GMM), is often used for model-based clustering because a single probability distribution can be interpreted as corresponding to a single cluster. The expectation-maximization (EM) algorithm~\cite{em_algorithm} is commonly used to estimate FMM. Like $K$-means, the user must set $K$, and multiple trials are required.

The traditional way to determine the number of clusters using a clustering method, such as $K$-means or EM algorithm, is to run starting from $K=K_{\rm min}$ and increasing $K$ one by one until $K=K_{\rm max}$, a candidate range of the number of clusters $[K_{\rm min}, K_{\rm max}]$, then evaluate each result using a cluster validity index or an information criterion. This approach is inefficient because the clustering results already performed, say the result of $K-1$, are not considered to obtain the result of $K$. Therefore, several methods have been proposed to find an appropriate $K$ in an exploratory manner~\cite{HANCER201749}.

A straightforward extension of $K$-means is repeatedly dividing a cluster into two based on some criteria. X-means~\cite{pelleg2000x} is the method to search clusters using BIC as the splitting criterion, and G-means~\cite{hamerly2004learning} uses a normality test statistic of a cluster for the splitting criterion. Dip-means~\cite{kalogeratos2012dip} has also been proposed using an unimodality test statistic for the splitting criterion. In contrast, the basic idea in the model-based approach is to perform stepwise model selection. MML-EM~\cite{figueiredo2002unsupervised} searches for the appropriate model order by minimum message length (MML)~\cite{MML}, starting from a sufficiently large $K$ and eliminating components with small mixing probabilities. PG-means~\cite{feng2007pg} explores the suitable model by starting with a small $K$ and increasing it until the goodness-of-fit test statistics are no longer significant. Like MML-EM, the shrinking maximum likelihood self-organizing map (SMLSOM)~\cite{smlsom} starts from a sufficiently large $K$, merging unnecessary clusters into others. Unlike MML-EM, SMLSOM does not eliminate clusters with few members but actively creates similar clusters and merges them to search for the number of clusters. In this study, we call the algorithms mentioned above {\it cluster search} algorithms.

Evaluation of statistical methods by simulation is important, and systematic experimentation to characterize the methods is strongly recommended \cite{morris2019}. Numerous simulation studies have been conducted in clustering (for example, \cite{mechelen2023} introduces some recent works). Methodologies for comparative clustering studies by simulation have also been discussed~\cite{hennig2018}\cite{mechelen2023}. For cluster search algorithms, however, as \cite{HANCER201749} points out, there are few systematic evaluations of the algorithms described above, even if limited to centroid-based methods. For example, a technical report~\cite{bies2009} is one of the few studies that compared X-means and G-means by simulation. However, the data generation model is too simple, using only spherical Gaussians with equal variance, and a uniform distribution. The evaluation is also limited; they only use the percentage of correct answers to the number of clusters by methods, so they do not evaluate sample affiliations.

Therefore, this study aims to clarify which cluster search algorithms work well in which cases through systematic artificial data experiments based on various cases that GMM can generate. The cases are considered by combining five factors: dimensionality, sample size, the number of clusters, cluster overlap, and covariance type. Higher-order interactions are interesting, but the enormous number of combinations makes them challenging to analyze. Therefore, we use Group Lasso~\cite{glasso} to select interactions. This method identifies combinations of factors that are relatively more effective than others.

The rest of this paper is organized as follows. Section 2 provides the data generation method and algorithms to be compared. Section 3 describes the experimental design and the analysis method. Section 4 presents the experimental results, including the reasons for the failure case of each algorithm. Section 5 summarizes the characteristics of each algorithm. Finally, Section 6 concludes the paper.
\section{Data model and Algorithms}
\subsection{Data model}
In this study, we used GMMs to generate data sets. Let $\bm{x} = (x_{1}, x_{2}, \ldots, x_{p})^t$ be a $p$-dimensional sample. Let $\bm{x}$ follow the GMM and denote its probability density function as 
\begin{equation}
f(\bm{x}\mid \bm{{\rm \Psi}}_K) = \sum_{k=1}^K \pi_k f(\bm{x} \mid \bm{\mu}_k,\,\bm{\Sigma}_k),
\end{equation}
where $\pi_1,\,\pi_2,\,\ldots,\,\pi_K$ are mixing probabilities that satisfy $\pi_k > 0$ and $\sum_{k=1}^K \pi_k=1$, $\bm{\mu}_k$ and $\bm{\Sigma}_k$ are mean vector and covariance matrix of $k$-th component, respectively. $\bm{{\rm \Psi}}_K=\{\bm{\mu}_1,\,\ldots,\,\bm{\mu}_K,\,\bm{\Sigma}_1,\,\ldots,\,\bm{\Sigma}_K,\,\pi_1,\,\ldots,\pi_{K-1}\}$ is the set of parameters of the model. $f(\cdot\mid \bm{\mu}_k,\,\bm{\Sigma}_k)$ is the Gaussian density function of $k$-th component, denoted as follows:
\begin{eqnarray}
 \lefteqn{f(\bm{x}\mid \bm{\mu}_k,\,\bm{\Sigma}_k)=} \nonumber \\
  &&
  \frac{1}{(2\pi)^{p/2}|\bm{\Sigma}_k|^{1/2}}\exp
  \left\{-\frac{1}{2}(\bm{x}-\bm{\mu}_k)^t\bm{\Sigma}_k ^{-1}(\bm{x}-\bm{\mu}_k)\right\}.
\end{eqnarray}
\subsection{Cluster search algorithm}
\label{sec: algorithms}
Let $\bm{x}_i = (x_{i1},\,x_{i2},\,\ldots,\,x_{ip})^t$ be the $i$-th sample of $n$ samples. Let $K$ be the number of clusters and $\{\bm{S}_k\}_{k=1}^K$ be a partition of a sample index set $\{1,\,2,\,\ldots,\,n\}$, in which each subset corresponds to samples belonging to each cluster. The number of samples in each cluster is $n_k=|\bm{S}_k|$. Note that this paper may use $\bm{S}\subseteq \{1,\,2,\,\ldots,\,n\}$ without a subscript to denote samples belonging to an arbitrary single cluster.
\subsubsection{X-means} 
X-means~\cite{pelleg2000x} searches for the number of clusters by splitting clusters. Clusters are represented by a single Gaussian model or a two-component GMM, and the BIC determines which model is adopted. X-means considers the following single Gaussian model $\mathcal{M}_1$ and two-component GMM $\mathcal{M}_2$ for the sample subset $\bm{S}$ and its two component partition $\{\bm{S}_1,\,\bm{S}_2\}$, respectively.
\begin{center}
\begin{tabular}{lcl}
$\mathcal{M}_1$ & : & $l_1=\log \prod_{i\in \bm{S}} f(\bm{x}_i\mid \bm{\mu},\,\sigma_1^2\bm{I})$, \\

$\mathcal{M}_2$ & : & $l_2=\log \left[\prod_{i\in \bm{S}_1} w f(\bm{x}_i\mid \bm{\mu}_1,\,\sigma_2^2\bm{I})\right] \left[\prod_{i\in \bm{S}_2} (1-w) f(\bm{x}_i\mid \bm{\mu}_2,\,\sigma_2^2\bm{I})\right]$,
\end{tabular}
\end{center}
where $0 < w < 1$ and $\bm{I}$ is the identity matrix of size $p$.
X-means compares $\mathcal{M}_1$ and $\mathcal{M}_2$ by BIC based on the maximum log-likelihood of each model to decide on splitting the cluster. The parameters of $\mathcal{M}_2$ are estimated from $\bm{S}_1$ and $\bm{S}_2$ obtained by spliting $\bm{S}$ using $K$-means. The proposal paper~\cite{hamerly2004learning} states that when dividing $\bm{S}$ by $K$-means, the initial two centroids are determined randomly, but the paper does not describe the specific method. In this study, the initialization was the same as the Dip-means method described later.

Due to the randomness of the initial values of $K$-means, it may be necessary to run X-means multiple times. In that case, X-means considers the following model to evaluate the fitness of the entire cluster to the data set for selecting the best result from multiple trials.
\begin{center}
\begin{tabular}{lcl}
$\mathcal{M}_K$ & : & $l_K=\sum_{k=1}^K \sum_{i\in \bm{S}_k}\log w_k f(\bm{x}_i\mid \bm{\mu}_k,\,\sigma_K^2 \bm{I})$.
\end{tabular}
\end{center}
The estimation of $\bm{\mu}_k$ is each cluster's centroid, and $0<w_k<1$ is obtained by $n_k/n$. Note $\sum_{k=1}^K w_k = 1$. Since the method for computing the estimation of $\sigma_K^2$ is not given in the proposal paper~\cite{hamerly2004learning}, we estimated it as $\frac{1}{n-K} \sum_{k=1}^K\sum_{i\in \bm{S}_k} \|\bm{x}_i-\bm{\mu}_k\|^2$ in this study. Note $\|\cdot\|$ is the $l_2$ norm.

X-means then evaluates the entire cluster using the following BIC formula.
\begin{equation}
\label{eq: bic_xmeans}
{\rm BIC_{Xmeans}} = -l_K + \frac{K(p+1)}{2} \log n.
\end{equation}

This BIC also can be applied to other centroid-based methods such as Dip-means. The recommended method to select the best result from multiple trials is not described in the proposal paper of Dip-means~\cite{kalogeratos2012dip}, so we used Eq.~(\ref{eq: bic_xmeans}) for Dip-means in this study.

\subsubsection{G-means}
G-means~\cite{hamerly2004learning}, like X-means, examines the division of individual clusters. The basic idea of G-means is to project the samples in a cluster into one dimension and consider a normality test for the projected samples under a significance level $\alpha$. G-means considers a scalar projection of $\bm{x}_i$ onto $p$-dimensional vector $\bm{P}$ as 
$x_i = \bm{P}^t \bm{x}_i,\ \forall i \in \bm{S}.$ G-means finds $\bm{P}$ in the following way. Like X-means, G-means obtains $\{\bm{S}_1,\,\bm{S}_2\}$ by dividing the sample subset $\bm{S}$ by $K$-means. Let $\bm{\mu}_1$ and $\bm{\mu}_2$ be centroids corresponding to $\bm{S}_1$ and $\bm{S}_2$, respectively. G-means defines $\bm{v} = \bm{\mu_1} - \bm{\mu}_2$ and calculates $\bm{P}=\bm{v} / \|\bm{v}\|$. The largest eigenvalue $\lambda$ and its eigenvector $\bm{u}$ of points in $\bm{S}$ are calculated to set $\bm{\mu} \pm \bm{u} \sqrt{2\lambda / \pi}$ as initial centroids, which the paper recommends.

G-means uses the Anderson--Daring (AD) test statistic of projected samples as a normality test statistic. $n$ projected samples $(x_i)_{i=1}^n$ are assumed to be standardized to a mean of 0 and variance of 1. In addition, projected samples are assumed to be ordered such that $x_1 < x_2 < \cdots < x_n$. Let $F$ be the cumulative distribution function of the standard normal distribution and $z_i=F(x_i)$; then, the (modified) AD statistic where $\mu$ and $\sigma^2$ are unknown is given by
\begin{equation}
A^2_* = A^2 (1 + 4/n - 25/n^2),
\label{ADstatistic}
\end{equation}
where $A^2 = -\frac{1}{n} \sum_{i=1}^n (2i-1)[\log(z_i) + \log(1-z_{n+1-i})] - n.$

The authors provide a critical value of 1.8692 corresponding to $\alpha = 0.0001$. Therefore, in G-means, the decision to divide cluster $k$ is made as follows. Obtain $A^2_*$ for $\bm{S}_k$ using the method mentioned above, and if $A^2_*>1.8692$, the division is adopted; otherwise $\bm{S}_k$ remains. The above procedure is repeated until no more clusters are added.
\subsubsection{Dip-means}
X-means and G-means search for the number of clusters by assuming a specific probability distribution for the data. Dip-means~\cite{kalogeratos2012dip} is a more flexible method that does not assume a specific distribution. The fundamental idea of Dip-means is to examine if the sample distribution within a cluster is unimodal and to divide the cluster if it is multimodal. Dip-means uses Hartigan's dip statistic~\cite{hartigan1985dip} to investigate unimodality.

Dip-means examines the unimodality of the distance distribution of points within a cluster. Euclidean distance is used here, but other distances can be used too. Dip-means considers each point of $\bm{S}_k$ as a {\it viewer} to detect the unimodality of cluster $k$. It calculates distances $\bm{d}_i = \{\|\bm{x}_i - \bm{x}_{i'}\| \mid i'\in\bm{S}_k\setminus\{i\}\}$ for all $i\in\bm{S}_k$ and tests the unimodality for each $\bm{d}_i$ with significance level $\alpha$. 
Note that the p-value of the unimodality test is calculated by the bootstrap method.
The significant samples are called {\it split viewers}. Let $v_k$ be the number of split viewers in $\bm{S}_k$. If the split viewer ratio $v_k/n_k$ is greater than the pre-determined threshold $v_{thd}$, then the cluster $\bm{S}_k$ is split into two clusters using $K$-means. Initial two centroids are chosen at a random point $\bm{x}\in\bm{S}_k$ and $\bm{\mu}_k - (\bm{x} - \bm{\mu}_k)$. Hyper-parameters of Dip-means are $b$, $\alpha$, and $v_{thd}$, where $b$ is the number of bootstrap sets. The authors use $b=1000$, $\alpha=0$ and $v_{thd} = 0.01$ in the paper~\cite{kalogeratos2012dip}.
\subsubsection{MML-EM}
Model selection for GMMs often involves estimating candidate models by the EM algorithm and then selecting the best result from the estimated candidates by BIC or other criteria. This method is computationally time-consuming because the EM is applied to each candidate model individually. In addition, the estimation of GMM with large $K$ by the EM algorithm sometimes estimates GMMs where some $\hat{\pi}_k$ is almost zero, making the estimates very unstable. To avoid these drawbacks, Figueiredo and Jain propose MML-EM~\cite{figueiredo2002unsupervised}, an efficient method to determine the number of components of GMM.

The primary idea of MML-EM~\cite{figueiredo2002unsupervised} is that estimation starts from a sufficiently large $K$ and components with small $\hat{\pi}_k$ are ``annihilated'', which means that the $\pi_k$ of the annihilated component is henceforth set to zero. MML-EM uses MML considering $\pi_k=0$ case as the objective function and component-wise EM algorithm (CEM2)~\cite{CEM2} to update each component successively.

Therefore, the algorithm repeats the convergence of CEM2 and the evaluation of MML until only a single component remains and returns the estimation result with the best MML evaluation.

\subsubsection{PG-means}
The fundamental idea of PG-means~\cite{feng2007pg} is similar to G-means, which also examines the distribution of projected samples. However, PG-means uses multiple projections instead of a single projection. Furthermore, it tests the goodness-of-fit of entire samples on the clusters instead of testing local samples. 
PG-means uses a GMM and estimates the model using the EM algorithm. The parameters of the model and samples are projected as
$
\mu_k = \bm{P}^t \bm{\mu}_k,\ \sigma_k^2 = \bm{P}^t \bm{\Sigma}_k \bm{P},\ k = 1,\,2,\,\ldots,\,K,
$ and
$
x_i = \bm{P}^t \bm{x}_i,\ i=1,\,2,\,\ldots,\,n.
$
Consider a one-dimensional GMM $\{\pi_k,\,\mu_k,\,\sigma^2_k\}_{k=1}^K$. The goodness-of-fit of the projected model on projected samples is investigated using the Kolmogorov--Smirnov (KS) test statistic. PG-means generates $h$ projections randomly according to $\bm{P}\sim N(0, 1/p \bm{I})$ and investigates each set of projected samples and model individually. Therefore, $h$ KS statistics are calculated to examine a cluster. If any KS statistic is larger than a threshold corresponding to significance level $\alpha$, PG-means adds a new cluster $K+1$. The parameters of a new cluster are set as follows: $\bm{\mu}_{K+1}$ is randomly chosen from entire samples, and $\bm{\Sigma}_{K+1}$ is the average of existing covariances. $\pi_{K+1}$ is set to $1/K$ and then all mixing probabilities are re-normalized. PG-means runs EM 10 times with $K+1$ clusters and finds the best likelihood to avoid a poor local solution. The above procedure is repeated until all KS statistics are no longer significant. Hyper-parameters of PG-means are $\alpha = 0.001$ and $h=12$.
\subsubsection{SMLSOM}
SMLSOM uses an extended version of Kohonen's self-organizing map (SOM)~\cite{kohonen} for constructing clusters. 
The extended SOM estimates the parameter of a probability distribution model by the method of moments. Like the online learning version of Kohonen's SOM, it uses the stochastic approximation method with $\tau_{\rm max}$ times iteration. SMLSOM measures the similarity between clusters based on Kullback-Leibler divergence~\cite{kullback1951information} and determines similarity using a threshold coefficient $\beta > 0$. 
Clusters considered similar are updated to be closer together in the SOM learning. Based on the MDL, the cluster is determined to be integrated into other clusters. Their SOM learning and merging clusters are repeated until the number of clusters is no longer reduced.

The hyperparameters for SMLSOM are threshold coefficient $\beta$, the number of iterations $\tau_{\rm max}$, and SOM's parameters. The authors recommend $\beta=15$, $\tau_{\rm max}=n$, and SOM's parameters are default values from the Kohonen\footnote{Kohonen: supervised and unsupervised SOMs, available at: \url{https://cran.r-project.org/web/packages/kohonen/index.html}} package in R.

\subsubsection{Computational complexity}
The computational complexity of each method is described here. X-means applies $K$-means $O(k)$ times to obtain $k$ clusters; $K$-means with $r$ iterations requires $O(Knpr)$ computations, while X-means uses only $K$-means for two centroids. Therefore, the computational complexity of X-means is at most $O(knpr)$ to obtain $k$ clusters. The same can be said for G-means and Dip-means, but these methods require extra computation for the cluster splitting criterion. G-means requires finding the maximum eigenvalue, which can be calculated efficiently using the power method, so it does not affect the overall computation. Dip-means also requires $O(bn\log n + n^2)$, where $b$ is the number of bootstrap samples, for calculating the distance matrix and the bootstrap method.
Assuming the number of iterations of the EM algorithm is at most $r$, PG-means, when terminated at $k$ clusters, requires $O(k^2np^2r + kn\log n)$ computations according to the authors' paper~\cite{feng2007pg}. Note that the inverse of the covariance matrix must be obtained when calculating likelihoods, but the computational complexity they describe does not include those inverse computations. Similarly, MML-EM requires $O(K_{\rm max}^2np^2 r)$ with $K_{\rm max}$ as the initial value. The computational complexity of SMLSOM with $\tau_{\rm max}=n$ is $O(K_{\rm max}^3 np^2)$ when repeated until the number of clusters reaches one~\cite{smlsom}. The actual computation times of each method will be examined in Section~\ref{sec: computational time}.

\section{Method}
This section describes an experimental procedure in this study. The experiment aims to determine which algorithm is appropriate for which scenario regarding the data generation mechanism. Table \ref{tab: exp_params} lists five factors of the data generation mechanism considered in this experiment. Note mixing probabilities were uniform in this study. Generating data for each combination of these factors and the clustering results from each algorithm is evaluated using the Adjusted Rand Index (ARI)~\cite{Hubert1985}. It is a six-factor experiment combining algorithms and five factors for data generation.

In this experiment, we are interested not only in the main effects of the factors but also in the interaction effects. Many factors addressed in this experiment are parameters involved in data generation, and the difficulty of clustering varies greatly depending on their combination. Therefore, it is expected that many interaction effects will be significant in the analysis of variance. It is difficult to interpret experimental results considering the large number of interaction effects. Therefore, we used Group Lasso~\cite{glasso} to select interactions with relatively large effects.

\begin{table}[t]
    \centering
    \caption{Parameters for data generation}
    \begin{tabular}{ll}
        \toprule
        \textbf{Factor} & \textbf{Level}\\ \midrule
        Sample size $n$ & 3000, 9000, (27000)\textsuperscript{\textit{1},\,\textit{2}}\\ 
        Dimensionality $p$ & 2, 6, 18\\ 
        Cluster overlap $\bar{\omega}$ &  0.01, 0.05, 0.1\\ 
        Number of Clusters $K^*$ & 3, 6, 12\\ 
        Covariance type & 
        1: Homogeneous and Spherical \\
        & \hspace{1em} $\bm{\Sigma}_1=\cdots=\bm{\Sigma}_{K^*}=\sigma^2 \bm{I},\ \sigma > 0$\\
        & 2: Homogeneous and Non-Spherical \\ 
        & \hspace{1em} $\bm{\Sigma}_1=\cdots=\bm{\Sigma}_{K^*}=\bm{\Sigma}\in \mathcal{R}^{p\times p}$\\
        & 3: Heterogeneous and Spherical \\
        & \hspace{1em} $\bm{\Sigma}_k=\sigma_k^2 \bm{I},\  \sigma_k > 0,\,k=1,\,2,\,\ldots,\,K^*$\\
        & 4: Heterogeneous and Non-Spherical \\
        & \hspace{1em} $\bm{\Sigma}_k\in \mathcal{R}^{p\times p},\quad k=1,\,2,\,\ldots,\,K^*$ \\
        \bottomrule
    \end{tabular}
    \label{tab: exp_params}
    \begin{minipage}{0.9\linewidth}
    \textsuperscript{\textit{1}} $n=27000$ case was only executed when $p=18$\\
    \textsuperscript{\textit{2}} Dip-means was not applied for $n=27000$ case due to the time consumption.
    \end{minipage}
\end{table}

\subsection{Experimental method}
\label{sec: experiment}
\subsubsection{Data generation}
\label{sec: data_generation}
MixSim package\footnote{MixSim: Simulating Data to Study Performance of Clustering Algorithms. Available at \url{https://cran.r-project.org/web/packages/MixSim/index.html}}~\cite{melnykov2012mixsim} can generate samples based on the GMM specifying cluster overlap. MixSim considers the following misclassification probability.
\begin{equation}
  \begin{array}{r}
   \omega _{l\mid k} ={\rm Pr}[\pi _k f(\bm{x}\mid
   \bm{\mu}_k,\,\bm{\Sigma}_k) < \pi _l f(\bm{x} \mid \bm{\mu}_l,
   \,\bm{\Sigma}_l)],\ l\neq k,\\
   {\rm where}\ \bm{x}\sim \mathcal{N}(\bm{\mu}_k,\,\bm{\Sigma}_k).
  \end{array}
\end{equation}
$\omega _{l\mid k}$ is the misclassification probability of a sample generated from the $k$-th component being mistakenly classified as the $l$-th component, and $\omega _{k\mid l}$ indicates vice versa. MixSim defines the overlap between two components as $\omega _{k\,l} = \omega _{k\mid l} + \omega _{l\mid k}$. We can specify $\bar{\omega}$, the average of $\omega_{k\,l}$.

We can generate data sets with a specific overlap level $\bar{\omega}$ using MixSim. Furthermore, MixSim can specify the four covariance types shown in Table~\ref{tab: exp_params}. The mean vector $\bm{\mu}_k$ is sampled from a $p$-dimensional uniform distribution. If the spherical structure is specified, $\bm{\Sigma}_k = \sigma_k^2 \bm{I}$, $\sigma_k$ is taken from the standard uniform distribution; if the non-spherical structure is specified, $\bm{\Sigma}_k$ is obtained from the standard Wishart distribution with parameter $p$ and $p+1$ degrees of freedom. If the homogeneous structure is specified, all covariances are the same; otherwise, all covariances differ.

\subsubsection{Evaluation index}
\label{sec: evaluation}
\begin{table}[t]
    \centering
    \caption{Two-way contingency table}
    \label{tab: ARI calculation}
    \begin{tabular}{clccc}
    \toprule
    & & \multicolumn{2}{c}{Partitioning $\mathcal{V}$} & \\ \cmidrule{3-4}
    & & Same & Different & Total\\ \cmidrule{2-4}
    \multirow{2}{*}{Partitioning $\mathcal{U}$} 
    & Same   & $T$ & $P-T$ & $P$\\ \cmidrule{2-4}
    & Different & $Q-T$ & $H-P-Q+T$ & $H-P$\\ \cmidrule{2-4}
    & Total & $Q$ & $H-Q$ & $H$\\
    \bottomrule
    \end{tabular}
    \begin{minipage}{0.85\textwidth}
        $T=\sum_{u=1}^U \sum_{v=1}^V\binom{n_{u\,v}}{2}$.\\
        $P=\sum_{u=1}^U \binom{n_{u\cdot}}{2}$, where $n_{u\cdot} = \sum_{v=1}^V n_{u\,v}$.\\
        $Q=\sum_{v=1}^V \binom{n_{\cdot v}}{2}$, where $n_{\cdot v} = \sum_{u=1}^U n_{u\,v}$.\\
        $H=\binom{n}{2}$.
    \end{minipage}
\end{table}
We used ARI as the evaluation index in this experiment and analyzed obtained ARIs by linear regression model using the factors. Since ARI is a quantity that takes values between 0 and 1, it is necessary to transform it when performing linear regression analysis. The probit transformed amount of ARI was used as the response variable. However, ARI is zero when an algorithm estimates only one cluster, which becomes negative infinity by probit transformation. Therefore, a slight modification was made to avoid the ARI being zero.

Let $\mathcal{U}=\{\bm{u}_1,\,\bm{u}_2,\,\ldots,\,\bm{u}_U\}$ and $\mathcal{V}=\{\bm{v}_1,\,\bm{v}_2,\,\ldots,\,\bm{v}_V \}$ be two partitions of the $n$ samples and $n_{u\,v}\,(u=1,\,2,\,\ldots,\,U,\,v=1,\,2,\,\ldots,\,V)$ be the number of samples that belong in common to $\bm{u}_u$ and $\bm{v}_v$. We can then compute a two-way table such as the Table~\ref{tab: ARI calculation} for two sample pairs chosen from $n$ samples. Using this binary table, ARI can be expressed as follows.
\begin{equation}
\label{eq: ARI}
    {\rm ARI} = \frac{T-E}{M-E},
\end{equation}
where $E=\frac{QP}{H}$ and $M=\frac{1}{2}(Q+P)$.

We used the following modification to prevent ARI from becoming zero.
\begin{equation}
\label{eq: cARI}
    {\rm cARI} = \frac{T-E + 0.5n}{M-E + n}.
\end{equation}
This modification is such that all possible values of the Eq.~(\ref{eq: ARI}) are slightly closer to 0.5. Of course, other corrections are possible, but here, we simply adopted Eq.~(\ref{eq: cARI}) as a correction that avoids setting the ARI to zero and does not significantly change the overall evaluation values. Our experimental results showed that the modification was at most $\pm 0.004$.

\subsubsection{Simulation procedure}
\label{sec: simulation}

\begin{table}[!ht]
\centering
\caption{The number of replications for each factor level}
\label{tab: replications}
\begin{tabular}{lc>{\centering}p{8em}>{\centering \arraybackslash}p{8em}}
\toprule
& \textbf{Overall} & \multicolumn{2}{c}{\small \textbf{Dip-means and the other five methods}}\\ \cmidrule{3-4}
\textbf{Factor} & & \textbf{Dip-means} & \textbf{Other } \\ 
\hspace{1em}\textbf{Levels} & $N = 1,476$ & $N = 216$ & $N = 252$\\
\midrule
$n$ &  &  &  \\ 
\cmidrule(l{1em}){1-4}
\hspace{1em}3000 & 648 (44\%) & 108 (50\%) & 108 (43\%) \\ 
\hspace{1em}9000 & 648 (44\%) & 108 (50\%) & 108 (43\%) \\
\hspace{1em}27000 & 180 (12\%) &  0 (0\%) & 36 (14\%) \\
\midrule
$p$ &  &  &  \\ 
\cmidrule(l{1em}){1-4}
\hspace{1em}2 & 432 (29\%) & 72 (33\%) & 72 (29\%) \\ 
\hspace{1em}6 & 432 (29\%) & 72 (33\%) & 72 (29\%) \\ 
\hspace{1em}18 & 612 (41\%) & 72 (33\%) & 108 (43\%) \\ \midrule
$\bar{\omega}$ &  &  &  \\
\cmidrule(l{1em}){1-4}
\hspace{1em}0.01 & 492 (33\%) & 72 (33\%) & 84 (33\%) \\ 
\hspace{1em}0.05 & 492 (33\%) & 72 (33\%) & 84 (33\%) \\ 
\hspace{1em}0.1 & 492 (33\%) & 72 (33\%) & 84 (33\%) \\
\midrule
$K^*$ &  &  &  \\ 
\cmidrule(l{1em}){1-4}
\hspace{1em}3 & 492 (33\%) & 72 (33\%) & 84 (33\%) \\ 
\hspace{1em}6 & 492 (33\%) & 72 (33\%) & 84 (33\%) \\ 
\hspace{1em}12 & 492 (33\%) & 72 (33\%) & 84 (33\%) \\ 
\midrule
Covariance type &  &  &  \\
\cmidrule(l{1em}){1-4}
\hspace{1em}1 & 369 (25\%) & 54 (25\%) & 63 (25\%) \\ 
\hspace{1em}2 & 369 (25\%) & 54 (25\%) & 63 (25\%) \\ 
\hspace{1em}3 & 369 (25\%) & 54 (25\%) & 63 (25\%) \\ 
\hspace{1em}4 & 369 (25\%) & 54 (25\%) & 63 (25\%) \\
\bottomrule
\end{tabular}
\begin{minipage}{\textwidth}
$N$ is the number of samples\\
\end{minipage}
\end{table}

\begin{table}[t]
    \centering
    \caption{Parameters of each algorithm. }
    \label{tab: alg_params}
    \begin{tabular}{ll} 
    \toprule
        Algorithm & Parameters \\ \midrule
        X-means & $K_{\rm max}=50$, $K_{\rm min}=1$ \\
        G-means & $K_{\rm max}=50$, $K_{\rm min}=1$, $\alpha = 10^{-4}$\\
        Dip-means & $K_{\rm max}=50$, $K_{\rm min}=1$, $\alpha = 10^{-16}$, $v_{thd}=0.01$\\
        PG-means & $K_{\rm max}=50$, $K_{\rm min}=1$, $\alpha = 10^{-3}$, $h = 12$, $\epsilon = 10^{-4}$\\
        MML-EM & $K_{\rm max} = 20$, $K_{\rm min}=1$, $\epsilon = 10^{-4}$\\
        SMLSOM & $K_{\rm max} = 20$, $K_{\rm min}=1$, $\beta = 15$\\ \bottomrule
        & \multicolumn{1}{r}{\footnotesize $K_{\rm max}$: maximum number of estimated clusters}\\
        & \multicolumn{1}{r}{\footnotesize $K_{\rm min}$: minimum number of estimated clusters}\\
        & \multicolumn{1}{r}{\footnotesize $\alpha$: significance level} \\
        & \multicolumn{1}{r}{\footnotesize $v_{thd}$: split viewer ratio (see text)} \\
        & \multicolumn{1}{r}{\footnotesize $h$: number of projections}\\
        & \multicolumn{1}{r}{\footnotesize $\epsilon$: relative convergence tolerance for the EM algorithm}\\
    \end{tabular}
\end{table}

The data generation scenarios are shown in Table~\ref{tab: exp_params}. Note, $n=27000$ was set as an additional level to be run only for $p=18$ out of concern that at $p=18$, a small sample size would make the estimation of the model-based methods unstable, and then we could not capture tendencies. Dip-means was run without the $n=27000$ case because it could not be run in practical time. Therefore, Dip-means was run under 216 scenarios, while the other five methods were run under 252 scenarios by adding the $n=27000$ case. Thus, the total number of combinations of scenarios and methods considered in this experiment is 1476. The number of replications for each level is shown in Table~\ref{tab: replications}.

The simulation steps are as follows. Performed for each of the 1476 cases. The parameter settings for each algorithm are shown in Table~\ref{tab: alg_params}.
\begin{enumerate}
    \item Generate 100 data sets using the method described in \ref{sec: data_generation}.
    \item Run the algorithm 10 times for each data set.
    \item Select the best result based on the likelihood from 10 runs and compute the Eq.~(\ref{eq: cARI}).
    \item Calculate an average of evaluation indices for the 100 data sets and use the average value as an evaluation for the combination of scenario and algorithm in question.
\end{enumerate}

We used the following method to select the best result from 10 runs in the third procedure of the above. Note that G-means is a deterministic method, and all runs on the same data set will yield the same result, so it is performed only once. Eq.~(\ref{eq: bic_xmeans}) was used for X-means and Dip-means. For MML-EM and SMLSOM, the information criterion used in each method was used for each. Note we used the standard BIC for PG-means since no specific recommendation is described in the proposed paper~\cite{feng2007pg}.

\subsection{Analysis method}
\label{sec: grpreg}
This section describes how to analyze the experimental results obtained by the procedure described in \ref{sec: simulation}. Since this study aims to investigate how each method behaves in any given scenario, the interaction effects between the method and each factor in Table~\ref{tab: exp_params} are of interest. However, examining all factor combinations, including higher-order interaction effects, is unrealistic. Therefore, selecting the factors and interactions to be analyzed is necessary. In this experiment, it is necessary to identify factors and interactions with a more significant effect than the others. Although analysis of variance is the standard method for multiway experiments, it does not say anything about whether the effect of the factor and interaction of interest is larger or smaller than the effect of the others since it only compares the variation due to the variable individually with the residual variation. Therefore, consider the following variable selection by Group Lasso~\cite{glasso}.

When considering $J$ factors, $N\times D_j$ matrix $\bm{X}_j,\,j=1,\,2,\,\ldots J$ is the part of the design matrix corresponding to the $j$-th factor. The entire design matrix is the $N\times D$ matrix $\bm{X}=(\bm{X}_1\, \bm{X}_2\, \cdots \bm{X}_J)$, where $D=\sum_{j=1}^J D_j$. Group Lasso minimizes the following residual sum of squares with the penalty term.
\begin{equation*}
    \min_{\bm{\beta}\in \mathcal{R}^D} \left[\|\bm{y} - \bm{X}\bm{\beta}\|^2 + \lambda \sum_{j=1}^J\sqrt{D_j}\|\bm{\beta}_j\| \right],
\end{equation*}
where $\|\cdot\|$ is the $l_2$ norm, $\bm{y}$ is an $N$-dimensional vector representing the response variable, $\bm{\beta}_j$ is a $D_j$-dimensional coefficient vector corresponding to the $j$-th factor, and $\bm{\beta}=(\bm{\beta}_1^t,\,\bm{\beta}_2^t,\,\ldots,\,\bm{\beta}_J^t)^t$ is the entire coefficient vector. In this case, $\lambda \geq 0$ is the tuning parameter that determines the strength of the penalty.

The objective function of Group Lasso shows that the effect of a factor is compared to the residual variation by the first term and other factor effects by the second term. The factor with a smaller effect than the other factors shrinks to $\hat{\bm{\beta}}_j=\bm{0}$. The method by Group Lasso is compatible with the objective of variable selection in this experiment. Therefore, based on the above discussion, Group Lasso is used for variable selection in this study.

The tuning parameter is selected using cross-validation and information criteria such as AIC and BIC. However, the cross-validation method and AIC evaluate the expected prediction error and are indicators for forecasting problems. Since this experiment is an estimation problem, BIC, a criterion for selecting the model with the maximum posterior probability from the model class $\mathcal{M}$, is more appropriate. However, BIC may make an unreasonable choice because it considers a uniform prior distribution over the model class $\mathcal{M}$. For example, if we consider linear models as $\mathcal{M}$ and let $M_m\subset \mathcal{M}$ be the set of models with $m$ parameters, then ${\rm Pr}(M_m) < {\rm Pr}(M_{m'})$ when $m<m'$, and thus more complex models receive larger prior probabilities. When considering variable selection that includes factor interactions, as in this case, models that include higher-order interactions are more likely to be selected, which may cause interpretability problems. The extended BIC (EBIC)~\cite{EBIC} modifies this problem of prior probability distribution and is expressed for linear regression models as follows.
\begin{equation}
\label{EBIC}
    {\rm EBIC} = N\log\left(\frac{\rm RSS}{N}\right) + m\log N + 2\log \binom{D}{m},
\end{equation}
where ${\rm RSS}$ is the residual sum of squares. Let $m$ be the number of nonzero elements in $\bm{\beta}$ estimated by Group Lasso, and the tuning parameter $\lambda$ can be selected using Eq.~(\ref{EBIC}). Therefore, we use EBIC to tune the penalty parameter of Group Lasso in this study.

We now describe how the results are presented. The submatrix $\bm{X}_j$ corresponding to the $j$-th factor in the design matrix $\bm{X}$ depends on the chosen contrast. Thus, $\bm{\beta}$ is a reexpressed parameter through the contrast, and a coefficient of the level specified by the contrast is omitted to avoid rank deficiency of the design matrix. When comparing levels of a factor, the original redundant representation is sometimes more convenient than $\bm{\beta}$. The redundant parameters can be obtained as follows. For example, let $\bm{C}_A$ and $\bm{C}_B$ be contrast matrices of factor A and factor B, respectively, and consider the model up to two-factor interaction effects. Let the block diagonal matrix $\bm{C}={\rm diag}(\bm{C}_A,\, \bm{C}_B,\, \bm{C}_B \otimes \bm{C}_A)$, the redundant parameter $\bm{\alpha}$ can be obtained as follows~\cite{venables}. Note that $\otimes$ denotes the Kronecker product.
\begin{equation}
\bm{\alpha} = \bm{C}\bm{\beta}.
\end{equation}
It becomes more complicated for models with more than three-factor interactions. Here, we followed the procedure of \verb|dummy.coef| function in R.

In this study, linear regression analysis is conducted using the transformed quantity of the evaluated value in Eq.~(\ref{eq: cARI}) through the probit function for the reason described in Section~\ref{sec: evaluation}. As for the model, we first consider the main effects of the factor (\verb|method|) that levels the six algorithms in Table~\ref{tab: alg_params} and the five factors related to data generation in Table~\ref{tab: exp_params}. Furthermore, we consider up to four-factor interactions limited to the interaction with \verb|method|. Therefore, consider $J=6 + \binom{5}{1} + \binom{5}{2} + \binom{5}{3} = 31$. EBIC is used to select the tuning parameter $\lambda$ of Group Lasso for these variables. This study displays the results using $\bm{\alpha}$. In addition, sum-to-zero contrast~\cite{venables} is used as the contrast of each factor as in the traditional analysis of variance.
\section{Results}
\label{sec: results}
\begin{table}[t]
\centering
\caption{Descriptive statistics for each method}
\label{tab: summary_result}
\begin{tabular}{lSSSSS}
\toprule
 & \text{Min} & \text{Q1} & \text{Median} & \text{Q3} & \text{Max} \\ 
\midrule\addlinespace[2.5pt]
\multicolumn{6}{l}{cARI} \\ 
\midrule\addlinespace[2.5pt]
X-means & \num{8.08e-5} & 0.183 & 0.453 & 0.648 & 0.958 \\ 
G-means & \num{9.401e-4} & 0.354 & 0.521 & 0.721 & 0.947 \\ 
Dip-means & \num{2.423e-4} & 0.002 & 0.309 & 0.647 & 0.959 \\ 
PG-means & \num{3.00e-2} & 0.347 & 0.554 & 0.797 & 0.971 \\ 
SMLSOM & \num{8.080e-5} & 0.244 & 0.504 & 0.733 & 0.969 \\ 
MML-EM & \num{7.265e-4} & 0.393 & 0.624 & 0.861 & 0.971 \\ 
\midrule\addlinespace[2.5pt]
\multicolumn{6}{l}{$(\hat{K} - K^*) / (K^*-1)$} \\ 
\midrule\addlinespace[2.5pt]
X-means & -1 & -0.626 & -0.007 & 0.322 & 3.205 \\ 
G-means & -0.999 & 0.14 & 0.543 & 2.118 & 16.975 \\ 
Dip-means & -1 & -0.998 & -0.557 & -0.152 & 0.125 \\ 
PG-means & -0.909 & -0.635 & -0.391 & -0.112 & 0.06 \\ 
SMLSOM & -1 & -0.729 & -0.365 & -0.006 & 4.45 \\ 
MML-EM & -1 & -0.541 & -0.19 & -0.005 & 0.015 \\ 
\bottomrule
\end{tabular}
\end{table}

\begin{figure}[p]
\hspace{-0.1\textwidth}
\begin{tabular}{c}
    \begin{subfigure}[b]{1.2\textwidth}
      \includegraphics[bb= 0 0 648 216, width=\textwidth]{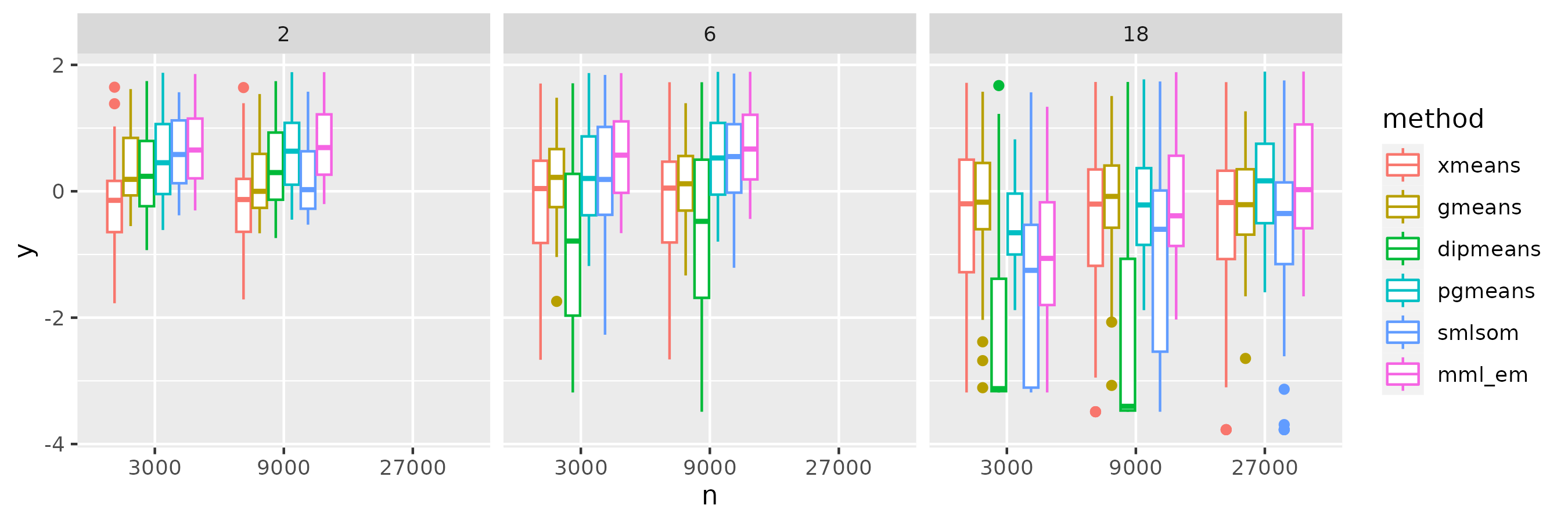}
      \caption{(method, $p$, $n$)}
      \label{fig: agg_case_n}
    \end{subfigure}\\
    \begin{subfigure}[b]{1.2\textwidth}
      \includegraphics[bb= 0 0 648 216, width=\textwidth]{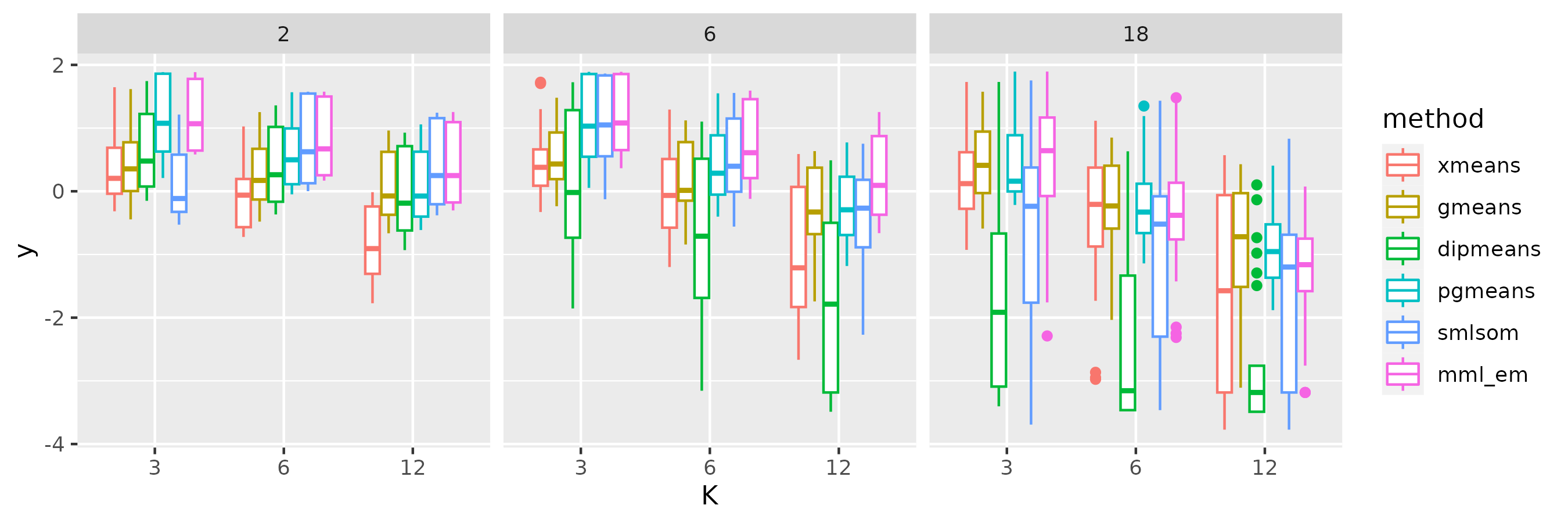}
      \caption{(method, $p$, $K^*$)}
      \label{fig: agg_case_K}
    \end{subfigure}\\
    \begin{subfigure}[b]{1.2\textwidth}
      \includegraphics[bb= 0 0 648 216, width=\textwidth]{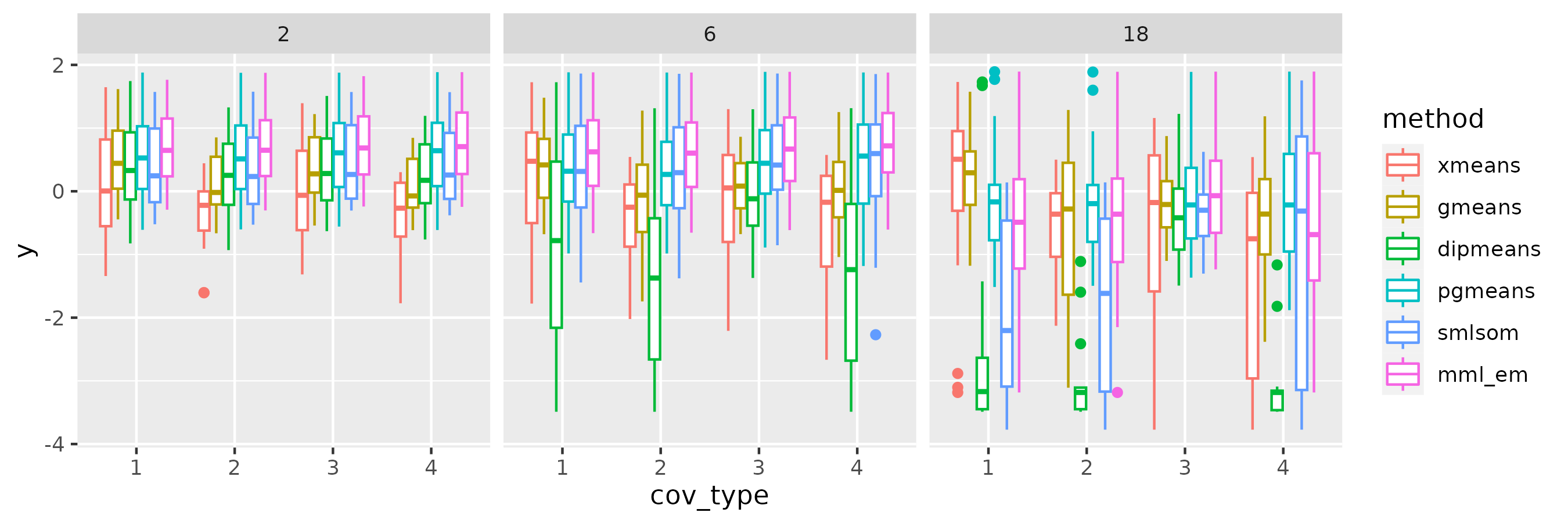}
      \caption{(method, $p$, covariance type)}
      \label{fig: agg_case_covtype}
    \end{subfigure}
\end{tabular}
\caption{Box plots of the response variable, which is the probit transformed cARI. Each figure corresponds to each factor combination selected as three-factor interactions.}
\label{fig: agg_plot}
\end{figure}

\begin{figure}[t]
  \hspace{-2.3cm}
  \begin{tabular}[b]{cc}
  \begin{tabular}[b]{c}
    \begin{subfigure}[b]{0.4\textwidth}
      \includegraphics[bb=0 0 225 225, width=\textwidth]{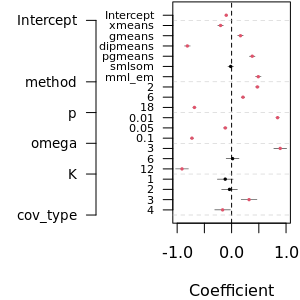}
      \caption{Main effects}
      \label{fig: main_effect}
    \end{subfigure}\\
    \begin{subfigure}[b]{0.4\textwidth}
      \includegraphics[bb=0 0 263 188, width=\textwidth]{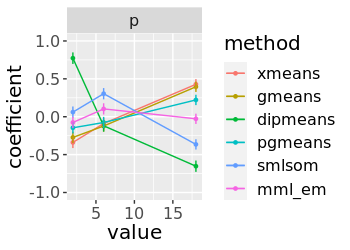}
      \caption{Two-factor interaction}
      \label{fig: intr_1st}
    \end{subfigure}
  \end{tabular}
  \begin{subfigure}[b]{0.9\textwidth}
    \includegraphics[bb=0 0 525 413, width=\textwidth]{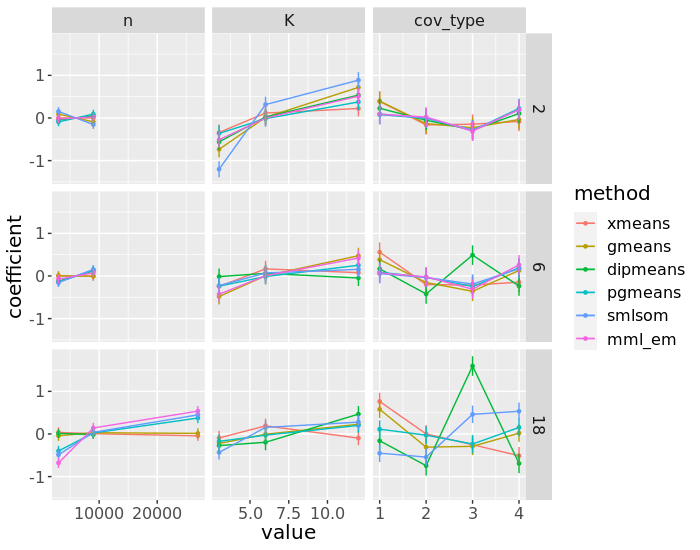}
    \caption{Three-factor interactions}
    \label{fig: intr_2nd}
  \end{subfigure}
  \end{tabular}
  \caption{Effect plot. (a) Main effects with confidence intervals. Black points indicate not significant. (b) Interaction effects between methods and $p$. (c) Three-factor interaction effects.}
  \label{fig: effects}
\end{figure}

\subsection{Descriptive results}
The descriptive statistics of the $N=1476$ evaluations obtained by the methods in Section~\ref{sec: simulation} are shown in Table~\ref{tab: summary_result} for each method. The ARI results show that all methods have a large range of values. Among them, MML-EM was superior on average. PG-means was the runner-up, but in 26 of the 252 scenarios, PG-means failed to select a model because the covariance matrix was singular in all 10 trials for some data sets. In this case, the results were evaluated, excluding data sets for which the model could not be estimated.

Table~\ref{tab: summary_result} also shows the evaluation of the cluster number estimate as $(\hat{K}-K^*)/(K^*-1)$. The $\hat{K}$ denotes the average of the cluster number estimates computed like the average of cARI obtained by Section~\ref{sec: simulation}. The difference from the true number of clusters is divided by $K^*-1$ to take the minimum value $-1$ when an algorithm outputs only one cluster. This measure indicates that G-means estimates the excess number of clusters compared to the true number.

\subsection{Analysis using Group Lasso}
Next, the factors and interactions were selected by the method described in Section~\ref{sec: grpreg}. We used \verb|grpreg|\footnote{grpreg: Regularization Paths for Regression Models with Grouped Covariates, available at: \url{https://cran.r-project.org/web/packages/grpreg/index.html}} package in R to run Group Lasso.  As a result, $J=12$ variables were selected among $J=31$ variables. For the main effect, all six factors were selected. Three two-factor interactions were selected: (method, $p$), (method, $K^*$), and (method, covariance type). Three three-factor interactions were selected: (method, $p$, $n$), (method, $p$, $K^*$), and (method, $p$, covariance type). The response variable distribution for the factor combinations selected as three-factor interactions is shown in box plots in Fig.~\ref{fig: agg_plot}. While this figure alone shows some tendencies, the following is a visualization of the estimated model to examine it in more detail. 

The coefficients were re-estimated using the ordinary least squares method for the model comprised of the selected factors and interactions. BIC for each model was 4532.7 for the $J=31$ model and $2581.4$ for the $J=12$ model, so the $J=12$ model was more applicable in terms of BIC. Note that the main effect of sample size $n$ among the above-selected factors was integrated into the three-factor interaction effect for interpretability. Also, for the same reason, the two-factor interactions regarding $K^*$ and covariance type were integrated into the three-factor interaction effects. Even with this expression change, RSS and its degrees of freedom are unchanged. The estimate of the redundancy parameter $\hat{\bm{\alpha}}$ is shown in Fig.~\ref{fig: effects}, which corresponds to the average value of each group of Fig.~\ref{fig: agg_plot} decomposed by the model.

In Fig.~\ref{fig: effects}, the values of $\hat{\bm{\alpha}}$ are indicated as points, and their 95\% confidence intervals are indicated by horizontal or vertical lines centered on the points. The variance of $\hat{\bm{\alpha}}$ was obtained as follows.
\begin{equation*}
    V(\hat{\bm{\alpha}}) = V(\bm{C}\hat{\bm{\beta}}) = \hat{\sigma}^2 \bm{C} (\bm{X}^t\bm{X})^{-1} \bm{C}^t,
\end{equation*}
where $\hat{\sigma}^2$ is the estimation of the error variance.

Fig.~\ref{fig: main_effect} shows the main effect, and the fundamental tendencies are as follows. As the number of dimensions $p$, the cluster overlap $\bar{\omega}$, and the number of true clusters $K^*$ become larger, the clustering problem becomes more difficult. The covariance type is not as influential as other factors related to data generation, but the problem is more difficult when the covariance is non-spherical and easier when the covariance is spherical. Among the methods, MML-EM and PG-means are superior to the other four methods. However, considering that PG-means was numerically less stable than MML-EM, we can say MML-EM is the better method among the six methods.

Fig.~\ref{fig: intr_1st} shows the two-factor interaction effects between methods and the number of dimensions $p$. The figure shows that X-means and G-means are advantageous over other methods in high dimensional scenarios with $p=18$. Looking at the case of $p=18$ in Fig.~\ref{fig: agg_plot}, there are cases where X-means and G-means were superior to the other methods. The centroid-based method may be advantageous because it estimates fewer parameters than the model-based method. On the other hand, the centroid-based method, Dip-means, did not work in most cases in high dimensions. The reason will be examined in the next section.

Fig.~\ref{fig: intr_2nd} shows the three-factor interaction effects. The interaction effect on sample size $n$ shows that increasing sample size $n$ improved the accuracy of model-based methods, especially in the high-dimensional case. In contrast, centroid-based methods such as X-means and G-means did not benefit from sample size increases. The interaction effect on the true number of clusters $K^*$ shows that there was not much difference in the starting point of the cluster number search, i.e., the approach of gradually increasing the number of clusters from a small number or gradually decreasing the number of clusters from a large number. The exception was SMLSOM, which shows a significant difference compared to the other methods in the case of $p=2$ at $K^*=3$, the furthest from the starting point $K_{\rm max}=20$. The interaction effect for the covariance type shows it was positive when the scenario matches the covariance structure assumed by the method. For example, this tendency was noticeable for X-means and SMLSOM. The difference between the methods was especially noticeable in the case of $p=18$. Heterogeneous and spherical covariance has a significant positive effect on Dip-means. We have seen that Dip-means does not work well for the $p=18$ case in the two-factor interaction effect, but this covariance type was an exception.

Two tendencies from the above results that are particularly striking and explainable from an algorithmic perspective are as follows. Each tendency is discussed in more detail in the next section.
\begin{itemize}
    \item G-means tends to significantly overestimate the number of clusters compared to other methods, as seen from Table~\ref{tab: summary_result}.
    \item Dip-means is significantly less accurate in high dimensions, regardless of the centroid-based method, which has fewer parameters to be estimated than the model-based method, as seen from Fig.~\ref {fig: intr_1st}.
\end{itemize}

\subsection{Investigation}
\label{sec: investigation}
\begin{figure}[t]
 \centering
 \includegraphics[bb=0 0 432 216, width=\textwidth]{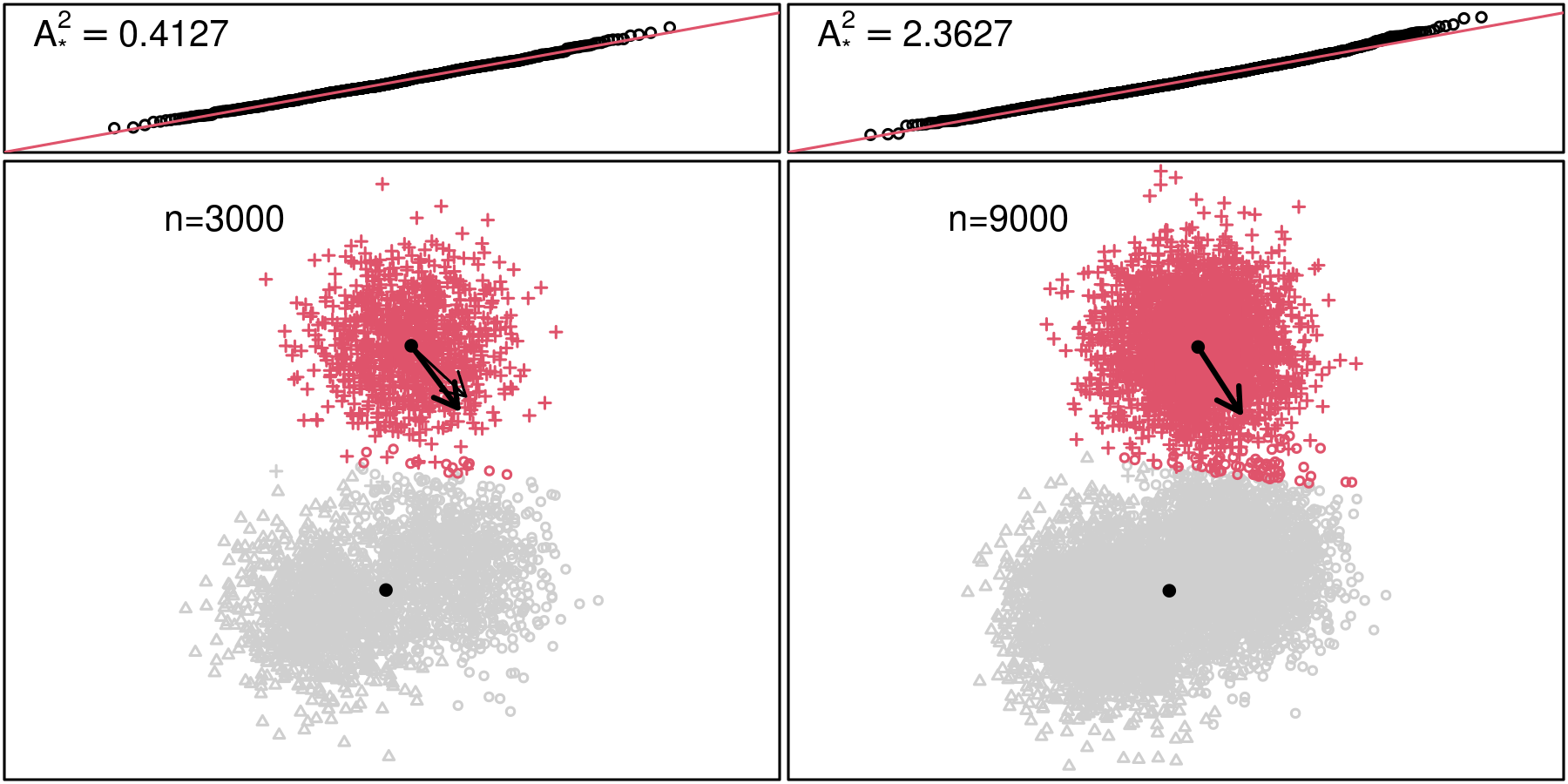}
 \caption{Example of the typical case that G-means does not work. The scatter plot shows two clusters obtained by $K$-means applied to the data set ($p=2$). The upper figure is the Q--Q plot of the red-colored cluster, calculated by G-means projection.}
 \label{gmeans_case}
\end{figure}

\begin{figure}[t]
 \centering
 \includegraphics[bb=0 0 576 144, width=\textwidth]{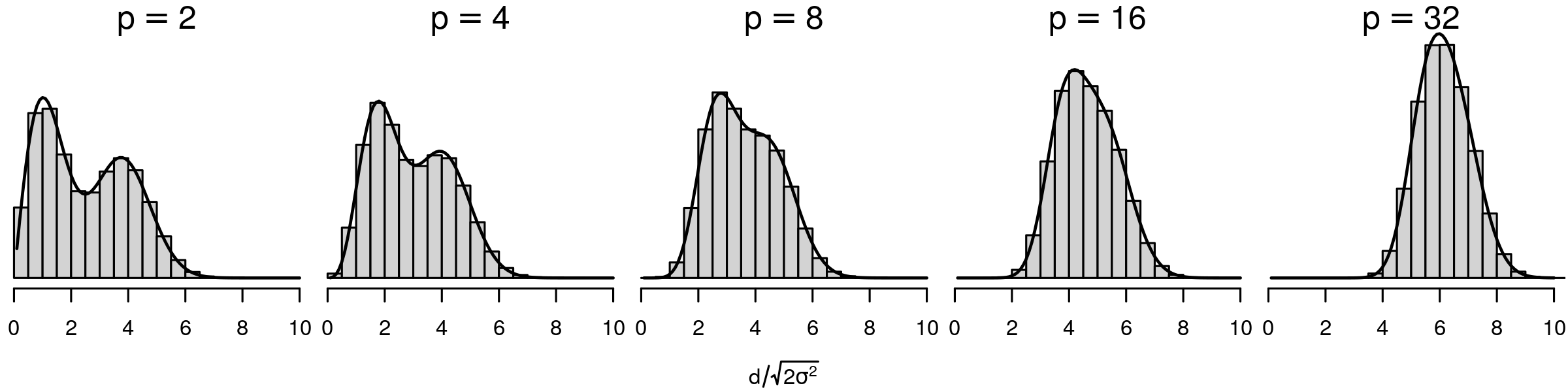}
 \caption{Histogram of Euclidean distances, calculated by $n=3000$ samples taken from the two-components GMM with spherical and homogeneous ($\bar{\omega} = 0.01$). The curve is the probability density function of a mixture of $\chi_p$ and $\chi_p(\lambda)$, where $\lambda = 3.4628$ in these cases.}
 \label{dipmeans_case}
\end{figure}
{\bf G-means} has the tendency of overestimating. A typical example of G-means failure is shown in Fig.~\ref{gmeans_case}. The figure shows the same $K^*=3$ homogeneous and spherical model ($\bar{\omega}=0.01$), where the data set differs only in the number of samples. Samples are already divided into red and gray clusters, and G-means tries splitting the red-colored cluster. The thin arrow is the direction $\bm{u}$ of the largest eigenvalue of the cluster, and the thick arrow is the projection direction $\bm{v}$ obtained by $K$-means. After projecting the samples in the red-colored cluster, $A_*^2$ can be calculated with Eq.~(\ref{ADstatistic}). The red-colored cluster covers one true cluster, and the gray-colored cluster covers two true clusters. In both cases, we can see that the red cluster has clipped samples that should have belonged to the gray cluster. As more samples are clipped, they skew the tail of the projected sample's distribution, making the AD test more likely to reject the current cluster. Therefore, $A_*^2$ was larger when $n=9000$ than when $n=3000$. In such cases, G-means will estimate an excessive number of clusters, which will cause the ARI to be worse.

{\bf Dip-means} does not work when $p$ are large. In the spherical and homogeneous ($\bar{\omega}=0.05$) case with $p=6$, as many as 53\% of the trials stopped at estimating only one cluster. This failure means many $\bm{d}_i$s are no longer multimodal.
The reason $\bm{d}_i$ is not multimodal in higher dimensions can be seen in Fig.~\ref{dipmeans_case}. Consider any two points $\bm{x}$, $ \bm{x}'$ in a two-component spherical and homogeneous GMM; when $\bm{x}$, $ \bm{x}'$ are identically distributed, $\|\bm{x}-\bm{x}'\|/ \sqrt{2\sigma^2}$ follows the chi distribution $\chi_p$ with $p$ degrees of freedom. For different distributions, it follows the non-central chi distribution $\chi_p(\lambda)$ with $p$ degrees of freedom and non-centrality $\lambda$, where $\lambda = \|\bm{\mu}_1 - \bm{\mu}_2\| / \sqrt{2\sigma^2}$. Therefore, $\|\bm{x}-\bm{x}'\| / \sqrt{2\sigma^2}$ of any two points $\bm{x}$, $ \bm{x}'$ follows a mixture of $\chi_p$ and $\chi_p(\lambda)$. If we fix $\bar{\omega}$, the only change by increasing $p$ is the degrees of freedom, while $\lambda$ is constant. Therefore, the $\|\bm{x}-\bm{x}'\|$ approaches a Gaussian-like distribution as $p$ increases and is no longer multimodal. Fig.~\ref{dipmeans_case} illustrates such a situation.

\subsection{Computational time}
\label{sec: computational time}
\begin{figure}[t]
    \hspace{-0.2\textwidth}
    \begin{tabular}{c}
        \includegraphics[bb=0 0 750 300, width=1.4\textwidth]{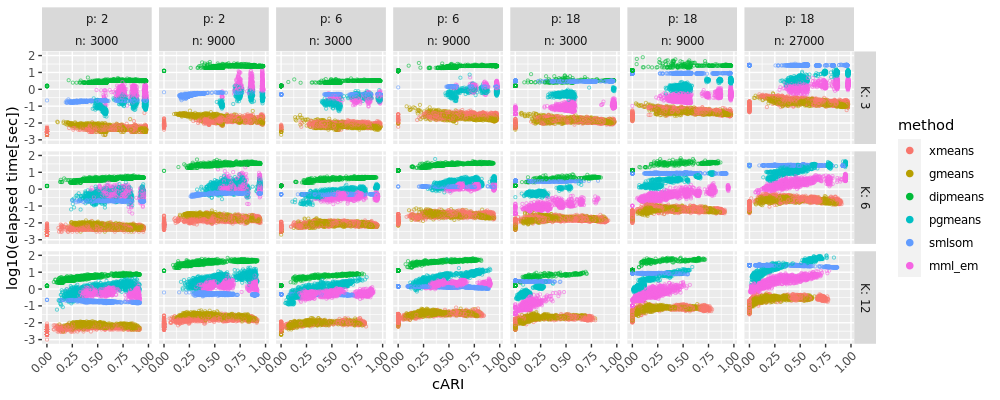}
    \end{tabular}
    \caption{Computation time and ARI. Each point represents the ARI of the result and the elapsed time taken by the algorithm to find the number of clusters for each data set. The representative value for each data set was chosen as described in Section~\ref{sec: simulation}.}
    \label{fig: computation_time}
\end{figure}

Fig.~\ref{fig: computation_time} compares each method regarding computation time and ARI. The figure shows that computation time and ARI are not necessarily a tradeoff. The figure also shows that X-means and G-means are equally the fastest among the six methods, and MML-EM is the fastest among the three model-based methods. In some cases, SMLSOM is faster than MML-EM, but SMLSOM drops in ranking as the number of dimensions $p$ increases, probably due to an implementation issue of SMLSOM\footnote{Unlike PG-means and MML-EM, SMLSOM updates the model one sample at a time. Learning one sample at a time can be inefficient when learning full covariance types because the covariance parameter of the probability density function used to determine sample affiliation varies from sample to sample.}. Dip-means is the same centroid-based method, such as X-means and G-means, but it takes the longest computation time among the six methods. Dip-means uses the distance between samples to calculate the dip statistic, so the distance matrix needs to be calculated in advance. The distance matrix calculation becomes a considerable computational burden when $n$ and $p$ are large.

Note we implemented each algorithm in C language. We ran all six algorithms on a Ubuntu 22.04 computer with two Xeon SC 4208 8C 2.1GHz and 128-GB memory.
\section{Discussion}
We examined each method's failure conditions in Section~\ref{sec: results}. Table~\ref{tab: discussion} summarizes in which cases each method fails. 

The centroid-based methods, X-means, G-means, and Dip-means, discover clusters by hard clustering with $K$-means. These methods have an advantage in computational time, except for Dip-means. However, clustering by Euclidean distance may be overly restrictive since it assumes the simple model, as seen in the description of X-means in Section~\ref{sec: algorithms}. Moreover, clusters may not be found well due to the nature of the Euclidean distance, as seen in Section~\ref{sec: investigation}.
In contrast, model-based methods such as PG-means and MML-EM work regardless of covariance type except for SMLSOM. However, in the case of full covariance, the number of parameters increases as $O(p^2)$, so without a sufficient sample size, these methods may not find clusters well due to the accuracy of the covariance estimation. In such cases, it is necessary to reduce the number of parameters to be estimated, such as by estimating the diagonal covariance structure.

Although each of the above six methods has drawbacks, Fig.~\ref{fig: computation_time} and Table~\ref{tab: discussion} suggest the choice of method.
MML-EM is the best choice among the six methods regarding performance and computation time. Although it has some problems with large $p$ small $n$, it is the most stable of the methods tested in this study for data sets of reasonable size. PG-means is the runner-up but has algorithmic instability that prevents it from obtaining estimates on some data sets and takes more computation time than MML-EM.

Therefore, based on the results of this study, MML-EM as the cluster number search method for GMM data sets should be given more attention.

\begin{table}[t]
    \caption{Summarization of failure conditions}
    \footnotesize
    \hspace{-0.1\textwidth}
    \begin{tabular}{llcccccc} \hline
    \toprule
    & & \multicolumn{3}{c}{Centroid-based algorithm} & \multicolumn{3}{c}{Model-based algorithm}\\ \cmidrule(lr){3-5} \cmidrule(lr){6-8}
    & & X-means & G-means & Dip-means & MML-EM & PG-means & SMLSOM\\ \midrule
    \multicolumn{2}{l}{Large $p$ with cluster overlap} & & & \xmark & & & \\
    \multicolumn{2}{l}{Large $n$ with cluster overlap} & & \xmark & & & & \\
    \multicolumn{2}{l}{Large $p$ small $n$} & & & & \xmark & \xmark &\xmark\\
    \multicolumn{2}{l}{$K^*$ far from start point} & & & & & &(\xmark)\textsuperscript{\textit{1}}\\
    \multicolumn{2}{l}{Covariance type:} & & & & & & \\
        & \multicolumn{1}{r}{Homogeneous} & & & & & &\xmark\\
        & \multicolumn{1}{r}{Heterogeneous} & \xmark & & & & &\\
    \bottomrule
    \end{tabular}
    \label{tab: discussion}
    \begin{minipage}{0.9\textwidth}
    \textsuperscript{\textit{1}} When $p$ is low.
    \end{minipage}
\end{table}
\section{Conclusion}
This study compared centroid- and model-based cluster search algorithms by various cases that GMM can generate. The cases considered were generated by combining four factors: covariance type, cluster overlap, the number of clusters, dimension, and sample size. Group Lasso was performed to select interactions and could provide a useful summarization for characterizing the methods. 

The results show that the cluster-splitting criteria of G-means and Dip-means can make unreasonable decisions in cases of large sample sizes or high dimensions when clusters overlap. The results also show that MML-EM, a model-based method, is effective for estimating the number of clusters in GMM data sets in terms of performance and computation time. We also found that the difference between the start point of the algorithms ($K_{\rm min}$ or $K_{\rm max}$) and the true number of clusters does not significantly affect the accuracy of the clustering when the number of dimensions is large.
%
%
Cases of imbalance in the number of samples per cluster have yet to be tested. This topic should be considered in future work.

\section*{Data availability}
Data and codes for the analyses performed in this study are available at \url{https://github.com/lipryou/cluster_search_simulation}

\section*{Funding}
This work was supported by JSPS KAKENHI Grant Number JP21H04600 and JST SPRING Grant Number JPMJSP2146.

\bibliographystyle{elsarticle-num}
\bibliography{main}

\begin{thebibliography}{10}
\expandafter\ifx\csname url\endcsname\relax
  \def\url#1{\texttt{#1}}\fi
\expandafter\ifx\csname urlprefix\endcsname\relax\def\urlprefix{URL }\fi
\expandafter\ifx\csname href\endcsname\relax
  \def\href#1#2{#2} \def\path#1{#1}\fi

\bibitem{centroid_based}
C.~Hennig, M.~Meila, {Cluster Analysis: An Overview}, in: C.~Hennig, M.~Meila,
  F.~Murtagh, R.~Rocci (Eds.), Handbook of Cluster Analysis, Chapman and
  Hall/CRC, 2015, Ch.~1, pp. 1--20.

\bibitem{bouveyron_celeux_murphy_raftery_2019}
C.~Bouveyron, G.~Celeux, T.~B. Murphy, A.~E. Raftery, Model-Based Clustering
  and Classification for Data Science: With Applications in R, Cambridge Series
  in Statistical and Probabilistic Mathematics, Cambridge University Press,
  2019.

\bibitem{EZUGWU2022104743}
A.~E. Ezugwu, A.~M. Ikotun, O.~O. Oyelade, L.~Abualigah, J.~O. Agushaka, C.~I.
  Eke, A.~A. Akinyelu, A comprehensive survey of clustering algorithms:
  State-of-the-art machine learning applications, taxonomy, challenges, and
  future research prospects, Engineering Applications of Artificial
  Intelligence 110 (2022) 104743.

\bibitem{nbclust}
M.~Charrad, N.~Ghazzali, V.~Boiteau, A.~Niknafs, {NbClust: an R package for
  determining the relevant number of clusters in a data set}, Journal of
  Statistical Software 61~(6) (2014) 1--36.

\bibitem{AIC}
H.~Akaike, A new look at the statistical model identification, IEEE
  Transactions on Automatic Control 19~(6) (1974) 716--723.

\bibitem{BIC}
G.~Schwarz, et~al., Estimating the dimension of a model, The Annals of
  Statistics 6~(2) (1978) 461--464.

\bibitem{MDL}
J.~Rissanen, Modeling by shortest data description, Automatica 14~(5) (1978)
  465--471.

\bibitem{em_algorithm}
A.~P. Dempster, N.~M. Laird, D.~B. Rubin, {Maximum likelihood from incomplete
  data via the EM algorithm}, Journal of the Royal Statistical Society. Series
  B (methodological) (1977) 1--38.

\bibitem{HANCER201749}
E.~Hancer, D.~Karaboga, A comprehensive survey of traditional, merge-split and
  evolutionary approaches proposed for determination of cluster number, Swarm
  and Evolutionary Computation 32 (2017) 49--67.

\bibitem{pelleg2000x}
D.~Pelleg, A.~Moore, X-means: Extending k-means with efficient estimation of
  the number of clusters., in: ICML, Vol.~1, 2000, pp. 727--734.

\bibitem{hamerly2004learning}
G.~Hamerly, C.~Elkan, Learning the k in k-means, in: Advances in neural
  information processing systems, 2004, pp. 281--288.

\bibitem{kalogeratos2012dip}
A.~Kalogeratos, A.~Likas, Dip-means: an incremental clustering method for
  estimating the number of clusters, in: Advances in neural information
  processing systems, 2012, pp. 2393--2401.

\bibitem{figueiredo2002unsupervised}
M.~A.~T. Figueiredo, A.~K. Jain, Unsupervised learning of finite mixture
  models, IEEE Transactions on Pattern Analysis and Machine Intelligence 24~(3)
  (2002) 381--396.

\bibitem{MML}
C.~S. Wallace, P.~R. Freeman, Estimation and inference by compact coding,
  Journal of the Royal Statistical Society. Series B (Methodological) (1987)
  240--265.

\bibitem{feng2007pg}
Y.~Feng, G.~Hamerly, {PG}-means: learning the number of clusters in data, in:
  Advances in neural information processing systems, 2007, pp. 393--400.

\bibitem{smlsom}
R.~Motegi, Y.~Seki, {SMLSOM: The shrinking maximum likelihood self-organizing
  map}, Computational Statistics \& Data Analysis 182 (2023) 107714.

\bibitem{morris2019}
T.~P. Morris, I.~R. White, M.~J. Crowther, Using simulation studies to evaluate
  statistical methods, Statistics in medicine 38~(11) (2019) 2074--2102.

\bibitem{mechelen2023}
I.~Van~Mechelen, A.~L. Boulesteix, R.~Dangl, N.~Dean, C.~Hennig, F.~Leisch,
  D.~Steinley, M.~J. Warrens, {A white paper on good research practices in
  benchmarking: The case of cluster analysis}, WIREs Data Mining and Knowledge
  Discovery 13~(6) (2023) e1511.

\bibitem{hennig2018}
C.~Hennig, Some thoughts on simulation studies to compare clustering methods,
  Archives of Data Science, Series A 5~(1) (2018) 1--21.

\bibitem{bies2009}
B.~Bies, K.~Dabbs, H.~Zou, On determining the number of clusters--a comparative
  study, Tech. rep., {Institute for Mathematics and Its Applications,
  University of Minnesota}, {IMA Preprints Series \#2486} (2009).

\bibitem{glasso}
M.~Yuan, Y.~Lin, Model selection and estimation in regression with grouped
  variables, Journal of the Royal Statistical Society Series B: Statistical
  Methodology 68~(1) (2006) 49--67.

\bibitem{hartigan1985dip}
J.~A. Hartigan, P.~Hartigan, The dip test of unimodality, The Annals of
  Statistics (1985) 70--84.

\bibitem{CEM2}
G.~Celeux, S.~Chretien, F.~Forbes, A.~Mkhadri, A component-wise {EM} algorithm
  for mixtures, Journal of Computational and Graphical Statistics 10~(4) (2001)
  697--712.

\bibitem{kohonen}
T.~Kohonen, {Self-Organizing Maps}, 3rd Edition, Springer, 2001.

\bibitem{kullback1951information}
S.~Kullback, R.~A. Leibler, On information and sufficiency, The Annals of
  Mathematical Statistics 22~(1) (1951) 79--86.

\bibitem{Hubert1985}
L.~Hubert, P.~Arabie, Comparing partitions, Journal of Classification 2~(1)
  (1985) 193--218.

\bibitem{melnykov2012mixsim}
V.~Melnykov, W.-C. Chen, R.~Maitra, {MixSim: An R package for simulating data
  to study performance of clustering algorithms}, Journal of Statistical
  Software 51~(12) (2012) 1--25.

\bibitem{EBIC}
J.~Chen, Z.~Chen, {Extended Bayesian information criteria for model selection
  with large model spaces}, Biometrika 95~(3) (2008) 759--771.

\bibitem{venables}
W.~N. Venables, B.~D. Ripley, {Modern applied statistics with S}, 4th Edition,
  Springer, 2002.

\end{thebibliography}

\newpage

\end{document}